\documentclass[lettersize,journal]{IEEEtran}
\usepackage{amsmath,amsfonts}
\usepackage{algorithmic}
\usepackage{algorithm}
\usepackage{array}
\usepackage[caption=false,font=normalsize,labelfont=sf,textfont=sf]{subfig}
\usepackage{textcomp}
\usepackage{stfloats}
\usepackage{url}
\usepackage{verbatim}
\usepackage{graphicx}
\usepackage{cite}
\usepackage{cuted}
\usepackage{capt-of}
\usepackage{booktabs}
\begin{document}
\bstctlcite{IEEEexample:BSTcontrol}
\title{BridgeACT: Bridging Human Demonstrations to Robot Actions via
Unified Tool-Target Affordances}

\author{
Yifan Han$^{1\ast}$,
Jianxiang Liu$^{2\ast}$,
Haoyu Zhang$^{3}$,
Yuqi Gu$^{2}$,
Yunhan Guo$^{2}$,
Wenzhao Lian$^{2\dagger}$%
\thanks{$^{1}$Institute of Automation, Chinese Academy of Sciences.;\;
$^{2}$School of Artificial Intelligence, Shanghai Jiao Tong University. \texttt{lianwenzhao@sjtu.edu.cn}.;\;
$^{3}$South China University of Technology.;\;
$^{\ast}$Equal contribution.;\;
$^{\dagger}$Corresponding Author.}%
}
\IEEEaftertitletext{%
\vspace{-1.2em}
\begin{center}
\includegraphics[width=0.95\textwidth]{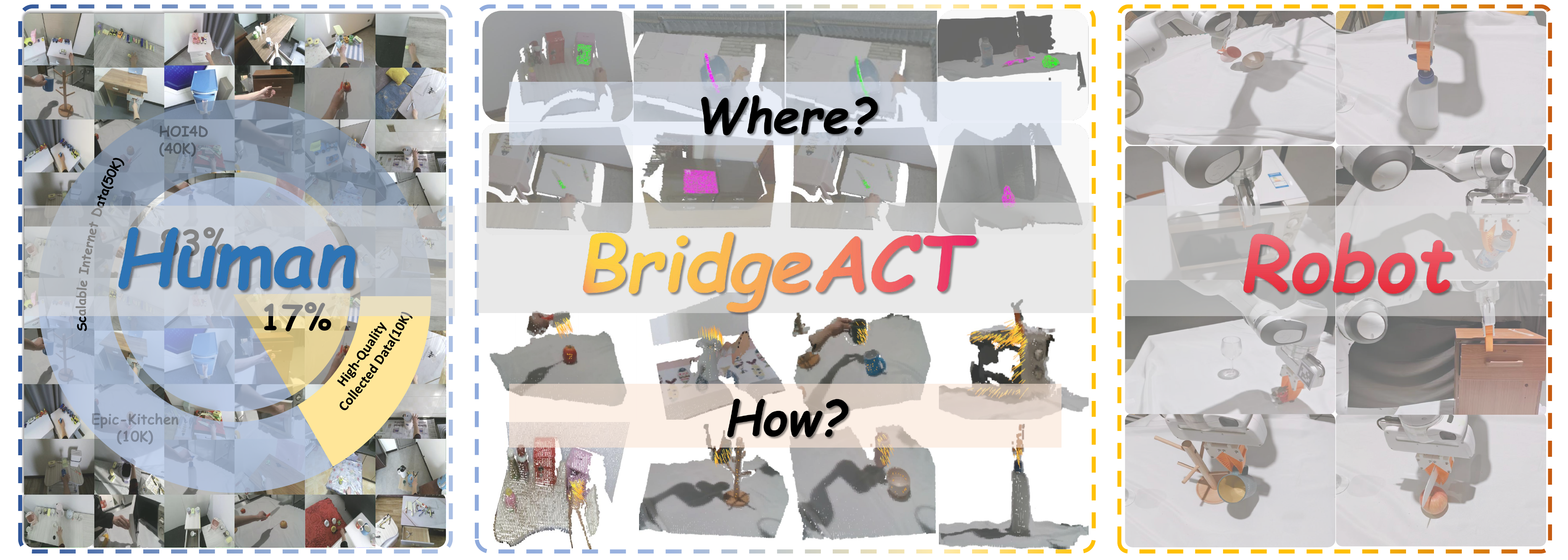}
\vspace{-0.8em}
\captionof{figure}{\textbf{Overview of BridgeACT.} BridgeACT learns role-conditioned tool-target affordances from human videos without robot demonstrations. It identifies task-relevant operable regions, assigns functional tool-target roles, and predicts executable 3D interaction dynamics for real-robot manipulation. The framework supports both single-object and object-to-object interactions across diverse manipulation scenarios.}
\label{fig:tizer}
\end{center}
}

\maketitle

\begin{abstract}
Learning robot manipulation from human videos is appealing due to the scale and diversity of human demonstrations, but transferring such demonstrations to executable robot behavior remains challenging. Prior work either relies on robot data for downstream adaptation or learns affordance representations that remain at the perception level and do not directly support real-world execution. We present BridgeACT, an affordance-driven framework that learns robotic manipulation directly from human videos without requiring any robot demonstration data. Our key idea is to model affordance as an embodiment-agnostic intermediate representation that bridges human demonstrations and robot actions. BridgeACT decomposes manipulation into two complementary problems: where to grasp and how to move. To this end, BridgeACT first grounds task-relevant affordance regions in the current scene, and then predicts task-conditioned 3D motion affordances from human demonstrations. The resulting affordances are mapped to robot actions through a grasping module and a lightweight closed-loop motion controller, enabling direct deployment on real robots. In addition, we represent complex manipulation tasks as compositions of affordance operations, which allows a unified treatment of diverse tasks and object-to-object interactions. Experiments on real-world manipulation tasks show that BridgeACT outperforms prior baselines and generalizes to unseen objects and scenes.

\end{abstract}


\section{Introduction}  

Human videos provide abundant observations of manipulation, but they are not robot demonstrations. Compared with robot data, human videos are easier to collect and contain richer variations in objects, scenes, viewpoints, and interaction patterns~\cite{liu2022hoi4d, damen2018scaling}. Recent works further show that motion priors extracted from human videos can provide useful guidance for robot execution~\cite{wen2023any, yuan2024general}. However, a human hand, a robot gripper, and a dexterous hand may execute the same task through different kinematics, contact strategies, and control spaces. Thus, directly transferring human motion or object trajectories to robot actions is fundamentally ill-posed.

The transferable signal in human demonstrations is not the absolute motion of a hand or an object, but the interaction structure that explains the task: which entity acts as the tool, which entity is affected as the target, where contact should occur, and how the tool should move relative to the target. This is particularly important in egocentric videos~\cite{liu2022hoi4d, damen2018scaling}, where camera motion is coupled with human head movement and makes absolute object trajectories unstable, while tool-target relations remain comparatively consistent. We therefore argue that human-to-robot transfer should be built upon role-conditioned interaction structures rather than embodiment-specific actions or object-centric motion alone.

Existing methods only partially capture this transferable structure. Many robot learning approaches still rely on robot demonstrations for policy training or downstream adaptation, which limits their scalability when learning from human videos~\cite{brohan2022rt, intelligence2025pi_, kim2024openvla, wen2023any}. Perception-level affordance methods can identify where interaction may occur, but the predicted visual cues are not directly executable as 3D robot motion~\cite{tang2025uad}. Flow-based methods predict object or point trajectories and provide more structured motion guidance~\cite{wen2023any, bharadhwaj2024track2act, yuan2024general}, but they are often object-centric or focus mainly on post-grasp motion, without explicitly modeling who acts on whom. VLM- or planner-based methods can generate symbolic task plans or keypoint constraints~\cite{huang2024rekep, wang2025vlm}, but they do not learn dense interaction dynamics from human demonstrations. As a result, existing approaches still lack a representation that jointly captures task-relevant contact regions, tool-target roles, and executable 3D motion.

To address this gap, we propose executable tool-target affordances as a role-conditioned 3D interaction representation for learning robot manipulation from human videos. Rather than parameterizing manipulation as human actions or object-centric trajectories, our representation describes a task through the functional relation between an acting tool and an affected target. For single-object manipulation, the tool corresponds to the human hand in demonstrations and the robot end-effector during deployment, while the target is the manipulated object. For object-to-object (O2O) manipulation, the tool denotes the interacting object and the target denotes the object being acted upon. This formulation abstracts away embodiment-specific action spaces while preserving the task-relevant contact structure and relative motion dynamics, providing a common interface for both single-object and O2O manipulation.

An executable tool-target affordance jointly encodes task-conditioned operable regions, functional tool-target roles, and future 3D interaction dynamics. These elements specify where interaction should occur, who acts on whom, and how the tool should move relative to the target. Unlike perception-level affordances that only localize possible interaction regions, the proposed representation is grounded on 3D query anchors and predicts their role-conditioned motion, making it directly compatible with downstream robot execution.

We instantiate this representation in a human-video-to-robot learning framework that requires no robot demonstration data. During training, the framework converts heterogeneous human videos into tool-target motion supervision through automatic task understanding, visual grounding, point tracking, and 3D trajectory reconstruction. During deployment, it grounds the relevant tool and target regions in a new robot scene, predicts role-conditioned 3D point flow with a diffusion-based motion affordance generator, and executes the predicted affordances through grasp selection and closed-loop SE(3) motion control. Real-world experiments on both single-object and O2O tasks show that the learned affordances can be instantiated as executable robot behaviors and generalize under object and scene changes.

In summary, we make the following contributions:
\begin{enumerate}
    \item We propose executable tool-target affordances, a role-conditioned 3D interaction representation that captures where to act, who acts on whom, and how the acting entity should move relative to the affected entity.

    \item We develop a human-video-to-affordance learning framework that automatically extracts task semantics, interaction entities, functional roles, and 3D motion supervision from heterogeneous human demonstrations, without requiring robot demonstration data.

    \item We instantiate the learned affordances on real robots through task-conditioned grounding, grasp selection, and closed-loop SE(3) motion control, demonstrating real-world execution and generalization across objects, scenes, and manipulation tasks.
\end{enumerate}

\section{Related Work}

\subsection{Learning Robot Manipulation from Human Demonstrations}

Learning robot manipulation from human demonstrations has become an increasingly important research direction. Compared with robot demonstrations, human videos are significantly cheaper to collect, more abundant at scale, and richer in scene diversity and interaction patterns. As a result, a growing body of work has explored how to reduce reliance on robot data by leveraging human demonstrations. For instance, methods such as~\cite{myers2024policy, tracegen, bharadhwaj2024track2act}  combine human and robot data during training to improve scalability and generalization. In contrast, works such as~\cite{wen2023any, niu2025pre} seek to first learn rich priors from human demonstrations, so that only a small amount of robot data is needed for downstream adaptation. More recently, with the rise of foundation models, approaches such as~\cite{wang2025vlm, huang2024rekep, myers2024policy} have begun to use large models to extract key execution signals directly from human demonstrations, such as subtask structure and task constraints, moving closer to robot learning frameworks that can operate from human videos alone.


\subsection{Structured 3D Flow Representations for Manipulation}

Representation learning has become a common and effective paradigm for learning manipulation from human videos. By extracting structured motion representations, these approaches aim to capture the most essential interaction dynamics from demonstrations while preserving useful physical priors, thereby providing a more interpretable and transferable interface for downstream control. Early works often focused on learning flow-like representations from 2D videos, such as~\cite{bharadhwaj2024track2act, wen2023any, xu2024flow}. However, purely 2D motion representations lack explicit spatial structure and are therefore limited in their ability to support embodied 3D manipulation. More recent work has accordingly shifted toward structured 3D flow representations, such as ~\cite{yuan2024general, niu2025pre, coil}, which better capture the spatial dynamics of object interaction and have shown improved performance. In parallel, a growing line of work on world-model representation learning, including ~\cite{pointworld}, seeks to learn abstract motion representations with explicit physical structure from robot videos, with the goal of improving scalability and generalization.

\section{Method}

Our goal is to learn a manipulation policy from human demonstrations that can be directly deployed on real robots. To this end, we propose BridgeACT, a framework that bridges human demonstrations and robot actions through affordance learning. Specifically, BridgeACT models affordance as an intermediate representation between objects and actions, enabling robots to learn both where to grasp and how to move. Unlike prior approaches that either rely on robot demonstration data or remain limited to perception-level affordance modeling, BridgeACT requires no robot data during training and instead learns executable manipulation policies for real-world deployment. This design further yields strong zero-shot generalization to unseen tasks and environments.

\subsection{Automatic Motion Point-Flow Dataset Construction from Human Demonstrations}

We train the Motion Affordance Generator using two sources of human demonstration data, including a large-scale public human-object interaction dataset~\cite{liu2022hoi4d,damen2020epic} and our own collected videos, without relying on any robot demonstration data. Specifically, we use six affordance categories from the public dataset, including \emph{open}, \emph{close}, \emph{pickup}, \emph{place}, \emph{push}, and \emph{pull}. Our collected data further supplements four additional affordances, namely \emph{pour}, \emph{press}, \emph{hang-on}, and \emph{cut}, resulting in a total of ten affordance types.
To unify different interaction patterns under a consistent supervision interface, we associate each sample with explicit \emph{tool} and \emph{target} roles. For single-object tasks, the target denotes the manipulated object, while the tool corresponds to the executor, i.e., the human hand during training and the robot gripper at test time. For object-to-object tasks, the tool denotes the interacting object, such as a knife, while the target denotes the affected object, such as a fruit. This role definition is used consistently throughout data construction and subsequent motion 3D-flow learning.

Importantly, we retain both tool and target queries in each sample, rather than modeling only the motion of the manipulated object. This design is motivated by the fact that our human-video sources include egocentric demonstrations, in which camera motion is coupled with human head movement. Under such viewpoint changes, the absolute motion of a single object in image space is not a stable supervision target. In contrast, especially for object-to-object interactions, the physically meaningful signal lies in the relative motion between the tool and the target, rather than the motion of either entity alone.

During preprocessing, we first segment the raw videos into fixed-length clips. Each clip spans 1.5 seconds and is downsampled to four frames within a sliding window, with a temporal stride of 0.5 seconds between adjacent clips. After this procedure, the resulting dataset contains more than 400 scenes, over 3{,}000 videos, and more than 60K clips. More importantly, this data construction pipeline is inherently scalable: it can continuously extract and curate training samples from large-scale human videos, and can in principle be extended to broader internet-scale video sources for affordance learning.

On this basis, we construct a unified three-stage data processing pipeline, consisting of VLM-based automatic task annotation, object-centric mask and trajectory extraction, and subsequent 3D reconstruction and filtering. The key principle is to convert heterogeneous human-video sources with different levels of annotation availability into a shared supervision format for motion affordance learning.

In the first stage, we use a vision-language model (VLM)~\cite{bai2025qwen3} for task understanding. Specifically, for each task, we uniformly sample 10 frames from a single video and feed them into the VLM to infer the task semantics. The VLM extracts both the manipulated entities and the action category involved in the task. We represent the manipulated entities in a unified tool-target format; when only a single object is involved, we use the human hand as the default executor to complete the role specification. The extracted objects and action are then organized into a unified textual template, which serves as the language input for the downstream model.

In the second stage, we construct 2D masks for the task-relevant objects. When object annotations are available, we directly use them as object masks. Otherwise, we employ SAM3~\cite{carion2025sam} to segment the objects of interest automatically. We then uniformly sample points within each mask and track them across frames using CoTracker3~\cite{karaev2025cotracker3}, yielding dense 2D trajectories for both the tool and target entities.

In the third stage, we construct 3D motion representations from the tracked object points. When reliable 3D annotations or object poses are available, we directly recover the corresponding 3D trajectories from these annotations. Otherwise, we lift the tracked 2D trajectories into 3D using estimated depth and camera geometry. In this way, data from different sources are converted into a unified 3D trajectory representation despite heterogeneous supervision conditions.

After 3D reconstruction, we apply a unified cleaning process to all samples, including approximating the human hand as a gripper-like structure, removing isolated point clusters and floating outliers, and retaining only the effective point set within the central interaction region. Finally, each processed sample is represented as
\[
\{P_{\text{scene}}, Q_{\text{tool}}, Q_{\text{target}}, l\},
\]
where $P_{\text{scene}}$ denotes the scene point cloud, $Q_{\text{tool}}$ and $Q_{\text{target}}$ denote the object-centric query point sets sampled from the tool and target point clouds, respectively, and $l$ denotes the language instruction.

\subsection{Task-Conditioned Affordance Grounding Agent}
\label{sec:aff_agent}


To instantiate the learned manipulation representation in the current scene, we design a lightweight task-conditioned affordance grounding module that converts abstract, broad, or underspecified task descriptions into localized affordance regions in the scene, rather than directly generating robot actions. The tool--target role definition here follows the data convention introduced earlier: for single-object tasks, the target denotes the manipulated object and the tool denotes the executor; for object-to-object (O2O) tasks, the tool and target denote the interacting tool object and the affected object, respectively. 

\textbf{Task understanding.} Given the current observation and task instruction, the module first uses an MLLM~\cite{bai2025qwen3} to normalize the raw instruction into a more explicit task formulation. 

\textbf{Affordance region understanding.} Based on the MLLM output, the module further parses the object description, the task-relevant affordance region description, and the prompt for visual segmentation, thereby compressing high-level task semantics into compact intermediate representations for subsequent visual grounding. 

\textbf{Prompt-conditioned segmentation.} The module then calls SAM3~\cite{carion2025sam} with the parsed prompt to segment candidate affordance regions in the scene, producing 2D masks $M_{\text{tool}}$ and $M_{\text{target}}$ for the tool and target regions, respectively. 

\textbf{Affordance verification.} Since the first-pass segmentation does not always accurately cover the true operable region, we further employ a vision--language verifier to assess whether $M_{\text{tool}}$ and $M_{\text{target}}$ are functionally valid for the intended interaction, rather than merely being visually consistent with the text description. 

\textbf{One-step recovery.} When the initial prediction is inaccurate but still recoverable, the module performs at most one additional recovery step using a more robust fallback prompt $\tilde{u}$ for re-grounding and segmentation. In practice, this recovery typically relaxes an unstable fine-grained part-level description into a more stable object-level or spatially broader region. The final verified masks, denoted as $\bar{M}_{tool}$ and $\bar{M}_{target}$, are projected onto the scene point cloud to obtain the grounded 3D query regions $Q_{tool}$ and $Q_{target}$, from which downstream motion queries are sampled.

\begin{figure*}[htbp]
  \centering
  \includegraphics[width=\textwidth]{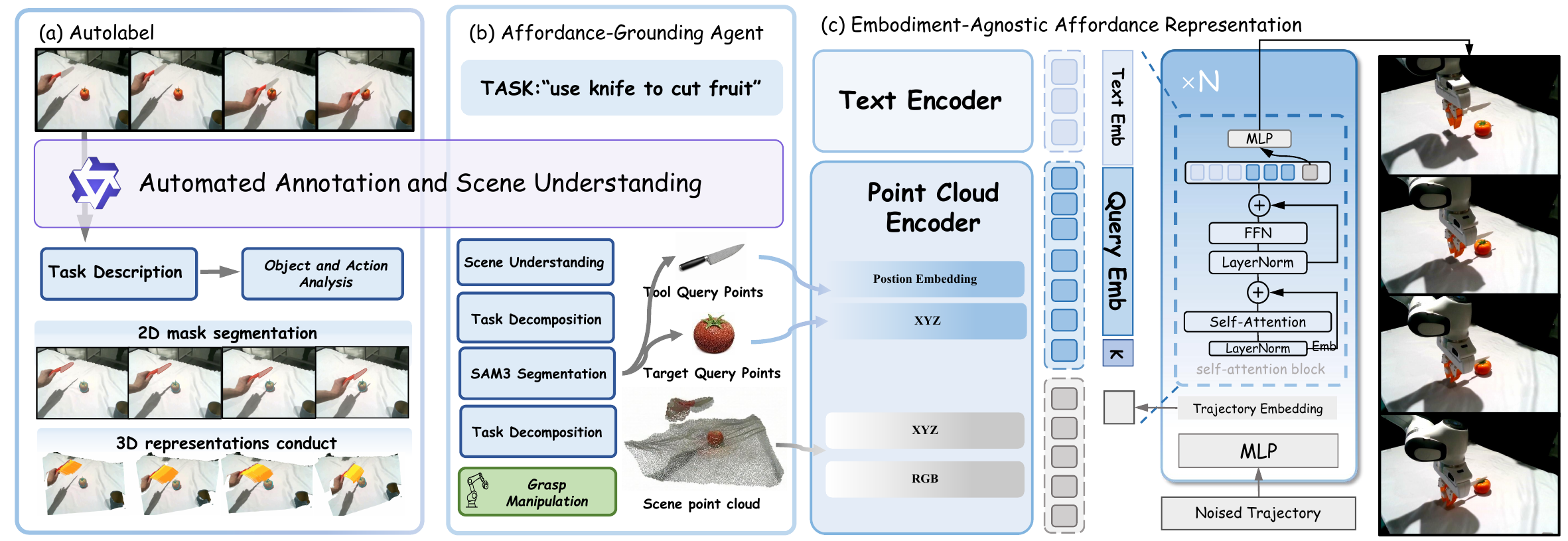}
  \caption{\textbf{Pipeline of BridgeACT.} 
(a) From raw human videos, we automatically construct motion point-flow training data through task understanding, object--action annotation, 2D segmentation, and 3D trajectory generation. 
(b) Given a task and the current scene, a task-conditioned affordance grounding agent localizes the task-relevant tool and target regions and samples 3D query points. 
(c) The grounded scene, query points, and language instruction are encoded into an embodiment-agnostic affordance representation, and a diffusion model predicts executable interaction-centric 3D point flow for robot manipulation without any robot demonstrations.}
  \label{fig:double}
\end{figure*}

\subsection{Embodiment-Agnostic Affordance Representation}

Each sample is represented as $O=\{P, Q_{\text{tool}}, Q_{\text{target}}, l\}$, where $P=\{(\mathbf{x}_j,\mathbf{c}_j)\}_{j=1}^{N_s}$ denotes the scene point cloud with 3D coordinates $\mathbf{x}_j\in\mathbb{R}^3$ and RGB values $\mathbf{c}_j\in\mathbb{R}^3$, $Q_{\text{tool}}=\{\mathbf{q}^{\text{tool}}_i\}_{i=1}^{N_t}$ and $Q_{\text{target}}=\{\mathbf{q}^{\text{target}}_i\}_{i=1}^{N_g}$ denote the grounded query point sets sampled from the tool and target regions, and $l$ denotes the language instruction. The tool and target queries are both retained throughout training and inference, rather than modeling only the motion of the manipulated object. This design is particularly important for human demonstrations containing egocentric videos, where camera motion is coupled with head movement and the absolute motion of a single object is not a stable supervision target. In particular, for object-to-object interactions, the physically meaningful signal is the relative motion between the tool and the target.

Unlike conventional point-cloud pipelines~\cite{wu2024point, qian2022pointnext} that encode query points only as latent conditions, we preserve the tool and target queries as explicit geometric anchors throughout the model. This design is motivated by two considerations. First, the query points are not only part of the input context, but also the prediction targets whose future positions define the motion affordance. Collapsing them prematurely into latent features weakens the pointwise correspondence between input anchors and predicted point flow. Second, our input is heterogeneous: scene points encode geometry and appearance, while tool and target queries encode only spatial anchors. The model therefore needs to fuse these different sources of information while preserving the functional role of the queries.

Concretely, scene points and query points are embedded separately. For a scene point $(\mathbf{x}_j,\mathbf{c}_j)\in P$, we use its 3D coordinate and RGB value as input attributes. For a query point $\mathbf{q}$ in $Q_{\text{tool}}\cup Q_{\text{target}}$, we use its 3D coordinate together with a positional embedding $\gamma(\mathbf{q})$, without RGB attributes. We then concatenate the scene tokens, tool-query tokens, and target-query tokens, and feed them into a PointNeXt~\cite{qian2022pointnext} encoder-decoder:
\[
\mathbf{H} = f_{\mathrm{dec}}(f_{\mathrm{enc}}(P, Q_{\text{tool}}, Q_{\text{target}})),
\]
where $\mathbf{H}$ denotes the pointwise features recovered at the original point resolution. From $\mathbf{H}$, we extract only the features corresponding to the query points, denoted by
\[
\mathbf{H}_{Q} = \mathbf{H}\!\mid_{Q_{\text{tool}}\cup Q_{\text{target}}}.
\]
These query-aligned features serve as the geometric condition for downstream motion prediction.

To inject task semantics, we encode the language instruction $l$ using CLIP~\cite{radford2021learning} and obtain a language feature $\mathbf{z}_l$. We then fuse $\mathbf{H}_{Q}$ with $\mathbf{z}_l$ through late fusion to obtain the final query condition
\[
\mathbf{C} = \mathrm{Fuse}(\mathbf{H}_{Q}, \mathbf{z}_l).
\]
Conditioned on $\mathbf{C}$, we apply a transformer-based diffusion model to predict the relative 3D displacement sequence for each query point. In implementation, the query condition is first projected into a single conditioning token, while the noisy displacement trajectory is reshaped into a sequence of step tokens. These tokens are concatenated as
\[
[\mathbf{c}_{\mathrm{cond}};\mathbf{x}_k^{1:m}],
\]
where $\mathbf{c}_{\mathrm{cond}}$ denotes the conditioning token derived from $\mathbf{C}$, and $\mathbf{x}_k^{1:m}$ denotes the noised displacement sequence at diffusion step $k$. The concatenated token sequence is then processed by a Transformer encoder~\cite{vaswani2017attention}, so that the displacement tokens interact with the conditioning token through self-attention. We therefore use prefix token conditioning, rather than explicit cross-attention, to inject the query-specific geometric and language information into the denoising process. The model outputs the predicted noise for each displacement step, from which the clean displacement sequence is recovered during sampling.

To train the diffusion module, we use a denoising noise-prediction objective. Let $\Delta \tilde{Q}^{1:m}(k)$ denote the noised ground-truth displacement sequence at diffusion step $k$, and let $\boldsymbol{\epsilon}\sim\mathcal{N}(\mathbf{0},\mathbf{I})$ denote Gaussian noise. We define diffusion loss as
\[
\mathcal{L}_{\text{diff}}
=
\mathbb{E}_{k,\boldsymbol{\epsilon}}
\Big[
\alpha(k)\,
\big\|
\boldsymbol{\epsilon}
-
\epsilon_{\theta}(\Delta \tilde{Q}^{1:m}(k),\, k,\, \mathbf{C})
\big\|_2^2
\Big],
\]
where $\alpha(k)$ denotes the min-SNR~\cite{hang2023efficient} reweighting factor at diffusion step $k$, which downweights timesteps with excessively large signal-to-noise ratios.

To further improve optimization and temporal stability, we introduce two additional designs. First, many query points are static or nearly static, especially those outside the effective interaction region. To reduce the bias toward trivial zero-motion samples, we introduce a motion-aware weighted loss that assigns larger weights to query points with larger ground-truth displacement magnitudes. Second, to mitigate drift caused by temporal integration, we further impose an accumulative displacement loss on the recovered trajectories, denoted as $\mathcal{L}_{\text{acc}}$. Specifically, let $\mathbf{s}_i^t=\mathbf{q}_i^t-\mathbf{q}_i^0$ and $\hat{\mathbf{s}}_i^t=\hat{\mathbf{q}}_i^t-\hat{\mathbf{q}}_i^0$ denote the ground-truth and predicted displacements relative to the initial frame, respectively. We further define
\[
r_i=\rho\!\left(\hat{\mathbf{s}}_i^{1:m}-\mathbf{s}_i^{1:m}\right), \qquad
w_i=g\!\left(\frac{1}{m}\sum_{t=1}^{m}\|\Delta \mathbf{q}_i^{t}\|\right),
\]
where $g(\cdot)$ is a monotonically increasing function and $\rho(\cdot)$ denotes a robust regression penalty. Then, the step loss and accumulative loss are defined as
\[
\begin{aligned}
\mathcal{L}_{\text{step}}
&=
\frac{1}{N_q}\sum_{i=1}^{N_q}\sum_{t=1}^{m}
\big\|
(\hat{\mathbf{q}}_i^{t}-\hat{\mathbf{q}}_i^{t-1})
-
\Delta \mathbf{q}_i^{t}
\big\|_2^2,\\
\mathcal{L}_{\text{acc}}
&=
(1-\lambda)\frac{1}{N_q}\sum_{i=1}^{N_q} r_i
+
\lambda\,
\frac{\sum_{i=1}^{N_q} w_i r_i}{\sum_{i=1}^{N_q} w_i},
\end{aligned}
\]
where $N_q=|Q_{\text{tool}}\cup Q_{\text{target}}|$.

The final objective is
\[
\mathcal{L}(\theta)=
\lambda_{\text{diff}}\mathcal{L}_{\text{diff}}
+
\lambda_{\text{step}}\mathcal{L}_{\text{step}}
+
\lambda_{\text{acc}}\mathcal{L}_{\text{acc}}.
\]

\subsection{Action Generation and Real-World Deployment}
The predicted affordances serve as an actionable intermediate representation that bridges perception and robot control. Since the affordance is represented in an explicit geometric form, it is naturally compatible with a wide range of downstream control and planning modules, including trajectory optimizers, model-based controllers, or other planners. For grasp affordance, we directly employ an off-the-shelf grasping module to determine a feasible pre-contact grasp pose.
 For motion affordance, we use a lightweight implicit policy to translate the predicted 3D point flow into robot actions. Specifically, we treat query points in the manipulated region as a local rigid set and estimate the rigid transformation that best aligns the predicted flow with the end-effector motion. This step is implemented as an ICP-style rigid registration problem and solved efficiently with SVD, recovering the corresponding rigid transformation in $SE(3)$ for the robot end-effector. During execution, the robot follows this transformation in a closed-loop manner, re-estimating the motion from updated observations at each step. In practice, we find that this simple mapping already provides reliable real-world execution and strong success rates.


\begin{figure*}[t]
    \centering
    \includegraphics[
        width=\textwidth,
        trim={0cm 30.5cm 47cm 0cm},
        clip
    ]{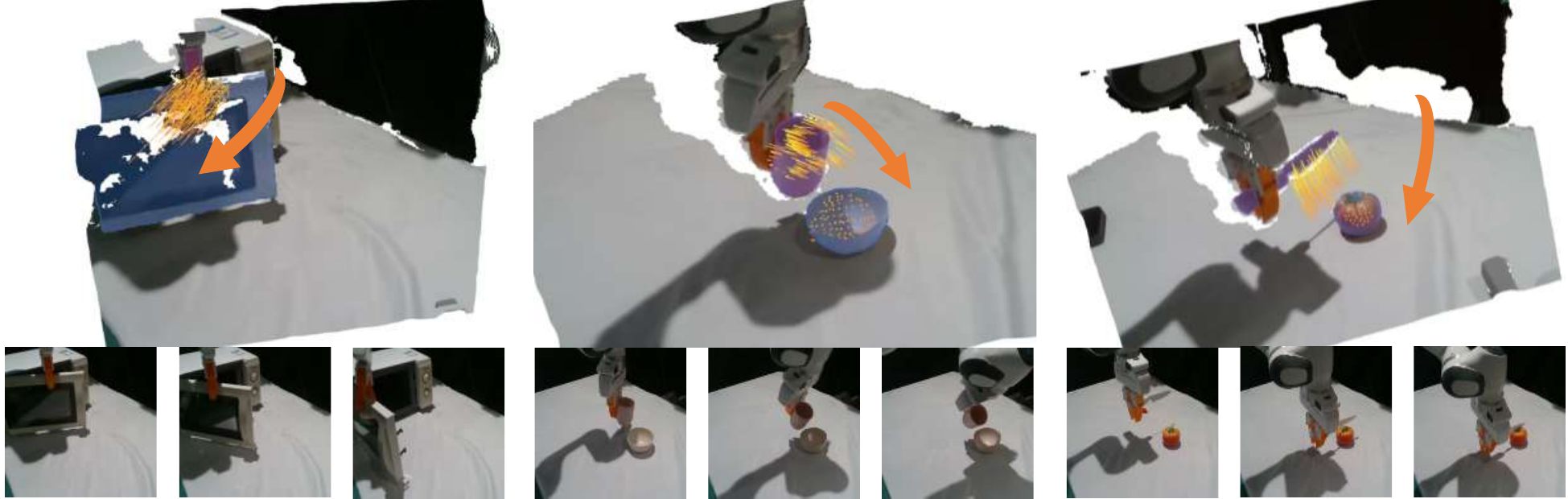}
    \vspace{-6mm}
    \captionsetup{font=normalsize}
    \caption{Partial visualization of motion affordances for representative tasks: open oven (left), pour water (middle), and cut fruit (right). The predicted 3D trajectories transfer effectively to real-robot execution across six affordance tasks.}
    \vspace{-4mm}
    \label{fig:lab}
\end{figure*}

\section{Experiment}

In this section, we conduct a series of experiments to answer the following core questions:

\textbf{Effectiveness of affordance representations.} Do the proposed grasp and motion affordances accurately capture the key interaction cues required for robotic manipulation, and can they effectively support the prediction of where to grasp and how to move?

\textbf{Bridging the human-to-robot gap.} Can affordance representations learned purely from human demonstrations transfer effectively to real-world robotic platforms, and does our method outperform existing baselines in both affordance prediction and task success rate?

\textbf{Zero-shot generalization skills.} Does our method maintain stable and reliable performance when deployed in unseen scenes, with novel objects and environments?

\textbf{Necessity of design choices.} How much does each proposed module and key design decision contribute to the overall performance, and are these components necessary for the gains achieved by our method?

\subsection{Affordance Representation Evaluation}

For motion affordance, we provide partial visualizations of our predicted motion affordances on a subset of tasks. As shown in Fig.~\ref{fig:lab}, our method produces meaningful 3D trajectories for tasks such as opening an oven, pouring water, and cutting fruit, and these trajectories transfer effectively to real-robot execution. Overall, successful transfer is observed across six affordance tasks.

We further compare our predicted 3D point flow against General Flow~\cite{yuan2024general}, which we select as the baseline because it likewise adopts a local point-level motion representation and directly maps the predicted flow to downstream robot execution, making it a direct and fair baseline for comparison. 

\begin{figure}[t]



\begin{minipage}[t]{0.48\columnwidth}
    \centering
    \includegraphics[height=3.2cm, keepaspectratio, trim={3cm 29cm 62cm 0cm}, clip]{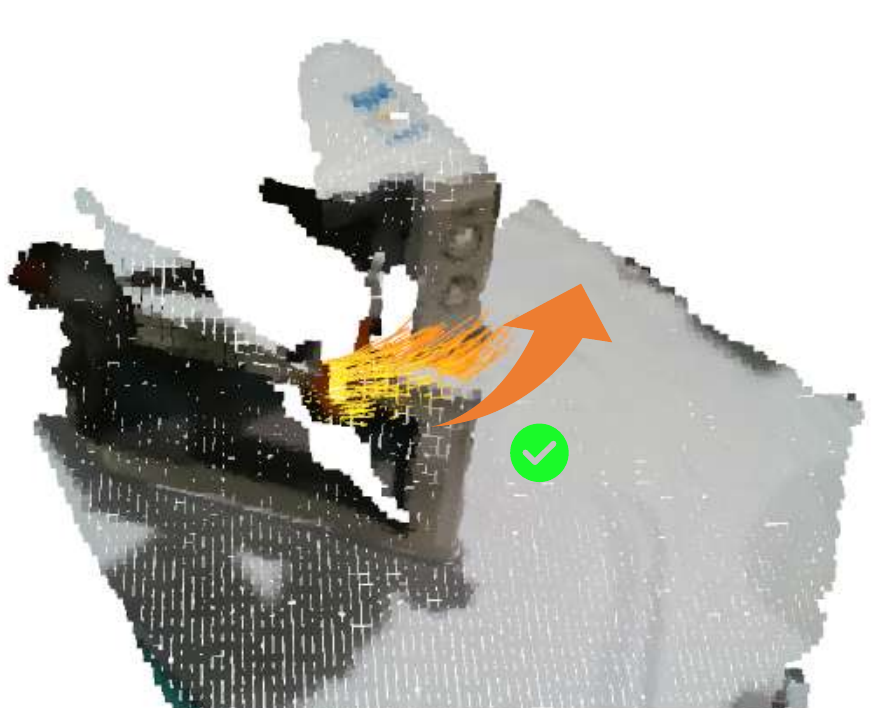}
    
    \vspace{0.5mm}
    \small (a)
\end{minipage}
\hfill
\begin{minipage}[t]{0.48\columnwidth}
    \centering
    \includegraphics[height=3.2cm, keepaspectratio, trim={23cm 29cm 41cm 0cm}, clip]{pic_s_2.pdf}
    
    \vspace{0.5mm}
    \small (b)
\end{minipage}

\vspace{1mm}

\begin{minipage}[t]{0.48\columnwidth}
    \centering
    \includegraphics[height=3.2cm, keepaspectratio, trim={1cm 13cm 63cm 16cm}, clip]{pic_s_2.pdf}
    
    \vspace{0.5mm}
    \small (c)
\end{minipage}
\hfill
\begin{minipage}[t]{0.48\columnwidth}
    \centering
    \includegraphics[height=3.2cm, keepaspectratio, trim={24cm 13cm 41cm 16cm}, clip]{pic_s_2.pdf}
    
    \vspace{0.5mm}
    \small (d)
\end{minipage}

\vspace{0mm}
\captionsetup{font=normalsize}
\caption{Trajectory visualization comparing our method with General Flow. The left column shows our results and the right column shows General Flow. The top row presents the in-domain task close oven, while the bottom row shows pickup basket with an out-of-domain object.}
\label{fig:ade_fde_qualitative}
\vspace{-7.8mm}
\end{figure}

As shown in Fig.~\ref{fig:ade_fde_qualitative}, for in-domain objects, our predicted flows are smoother and more stable, and more accurately capture the underlying motion geometry. For example, in the close action, General Flow exhibits an inverted spatial pattern, predicting longer trajectories on the inner side and shorter ones on the outer side, whereas our method better preserves the correct motion geometry, with longer trajectories on the outer side and shorter ones on the inner side. On out-of-domain objects, our method further shows stronger generalization. In the pickup action, General Flow even fails to recover the correct motion direction, producing lateral flows instead of the upward motion required for lifting, while our method remains consistent with the intended interaction semantics.
We attribute this improvement to two factors: our object-centric motion representation better captures task-relevant local dynamics, and our diffusion-based motion modeling offers a more expressive generative formulation than the CVAE-based design used in General Flow, leading to more stable predictions and better robustness under distribution shifts.

\subsection{Real-World Setup}

\begin{table}[t]
\centering
\caption{Results on six motion affordance types. Best results in \textbf{bold}.}
\label{tab:motion_affordance_results}
\renewcommand{\arraystretch}{1.25}
\setlength{\tabcolsep}{6pt}
\small
\begin{tabular}{lcccccc}
\toprule
Method & Pickup & Place & Open & Close & Pour & Cut \\
\midrule
ReKep        & 5/10 & 4/10 & - & - & 2/10 & 0/10 \\
Track2Act    & 4/10 & 1/10 & 0/10 & 0/10 & 0/10 & 0/10 \\
General Flow & 10/10 & 6/10 & 7/10 & 6/10 & 0/10 & 2/10 \\
\textbf{Ours} & \textbf{10/10} & \textbf{8/10} & \textbf{8/10} & \textbf{9/10} & \textbf{4/10} & \textbf{4/10} \\
\bottomrule
\end{tabular}
\vspace{-10pt}
\end{table}

For real-world experiments, we evaluate 6 tasks spanning six motion affordance types. Each task is paired with a corresponding grasp affordance, which specifies a feasible pre-contact grasp configuration and forms part of the overall task execution. The evaluated tasks include oven open/close, cup pickup/place, pouring water, and cutting fruit. 
During inference, our model enables closed-loop execution while allowing the number of query points to vary flexibly. Based on the affordance mask $M_{aff}$
provided by the Affordance Grounding Agent, we sample 128 query points from the tool and target regions, with a 3:1 allocation ratio.

For real-world evaluation, we mount an Intel RealSense D435 camera on the side of the robot and perform execution on a Franka manipulator equipped with a UMI gripper. During inference, we use RoboEngine~\cite{yuan2025roboengine} to segment out the robot arm. The Motion Generator is trained for 2,000 epochs, taking approximately one day on 8 NVIDIA H100 GPUs.

\textbf{Baselines}
We compare our method against representative baselines for real-world manipulation transfer.
1) \textbf{ReKep}~\cite{huang2024rekep} leverages large vision models and vision-language models to generate visually grounded manipulation specifications from free-form language instructions and RGB-D observations, and produces robot actions through a hierarchical optimization framework. We choose it as a relevant baseline because it also uses foundation-model-based scene understanding to bridge high-level task semantics and executable robot behavior.
2) \textbf{General Flow}~\cite{yuan2024general} predicts point trajectories via point-level flow guidance, but does not explicitly model the grasping stage. For a fair comparison, we assume that the target object has already been grasped by manually placing the gripper on the object and closing it before execution. Therefore, this baseline is evaluated starting from the post-contact manipulation stage.
3) \textbf{Track2Act}~\cite{bharadhwaj2024track2act} predicts flow affordances from 2D observations to guide execution. Since we do not use any robot demonstration data in our setting, we directly project its predicted flows to the robot for execution, without any additional fine-tuning or adaptation using robot data.

For baselines that require training, we train or fine-tune them on our data under the same supervision setting to ensure a fair comparison.

\subsection{Main Results and Analysis}

Table~\ref{tab:motion_affordance_results} shows the real-world experimental results. Overall, our method consistently outperforms the baselines across evaluation settings, demonstrating stronger executability and robustness in real-world manipulation. More importantly, these results show that our approach successfully bridges the human-to-robot gap: the robot is able to acquire and execute manipulation skills directly from human demonstration videos, without requiring any robot demonstration data. This highlights the effectiveness of affordance-based representations as a bridge between human demonstrations and robot execution.

The comparison also reveals characteristic limitations of the baselines. ReKep depends on explicitly selected visual keypoints, making its execution sensitive to keypoint localization and semantic correctness. In addition, its keypoint-based constraints often yield near-linear Cartesian motions, which are insufficient for articulated tasks such as opening and closing that require rotation-axis reasoning. General Flow does not explicitly model grasping, and therefore cannot cover the full manipulation pipeline. In addition, its CVAE-based generative design is less expressive for modeling complex, contact-rich motions, often resulting in larger trajectory deviations than our diffusion-based formulation. Track2Act, by contrast, uses 2D flow affordances to guide execution, but exhibits weaker robustness in preserving 3D geometric consistency. This issue is especially pronounced under partial occlusion or motions along the camera depth axis, where ambiguities in 2D-to-3D lifting can accumulate into larger execution errors and lower success rates.
Overall, these findings highlight the importance of explicitly modeling executable grasp and motion affordances in 3D, enabling more reliable and scalable transfer from human demonstrations to robotic manipulation.

\subsection{Zero-shot Skill Evaluation}





\begin{table}[t]
\centering
\caption{Generalization results under object and scene changes across three representative tasks: pick, open, and cut. Best results in \textbf{bold}.}
\label{tab:generalization_axes}
\renewcommand{\arraystretch}{1.25}
\setlength{\tabcolsep}{6pt}
\small
\begin{tabular}{lcccccc}
\toprule
& \multicolumn{3}{c}{Cross-Object} & \multicolumn{3}{c}{Cross-Scene} \\
\cmidrule(lr){2-4} \cmidrule(lr){5-7}
Method & Pick & Open & Cut & Pick & Open & Cut \\
\midrule
ReKep
& 5/10 & - & 2/10
& 4/10 & - & 1/10 \\

Track2Act
& 3/10 & 0/10 & 0/10
& 2/10 & 0/10 & 0/10 \\

General Flow
& 10/10 & 6/10 & 1/10
& 9/10 & 4/10 & 0/10 \\

\textbf{Ours}
& \textbf{10/10} & \textbf{7/10} & \textbf{3/10}
& \textbf{10/10} & \textbf{6/10} & \textbf{2/10} \\
\bottomrule
\end{tabular}
\vspace{-10pt}
\end{table}

\begin{figure}[t]
\begin{minipage}[t]{0.48\columnwidth}
    \centering
    \includegraphics[height=3.2cm, keepaspectratio, trim={6cm 29cm 62cm 3cm}, clip]{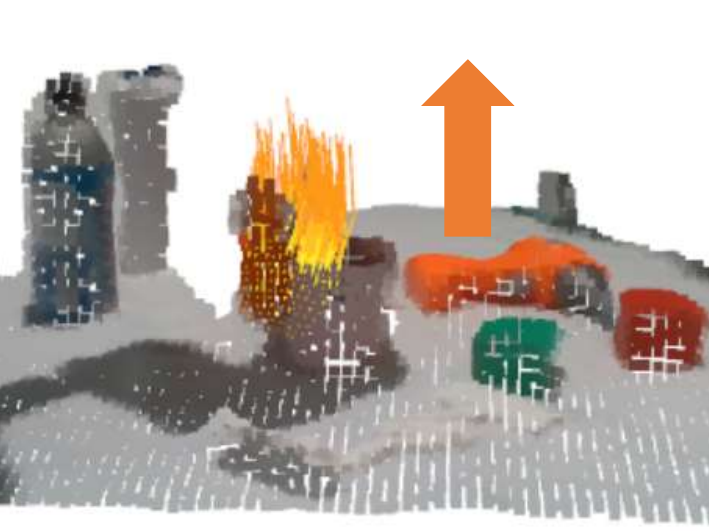}
    
    \vspace{0.5mm}
    \small (a)
\end{minipage}
\hfill
\begin{minipage}[t]{0.48\columnwidth}
    \centering
    \includegraphics[height=3.2cm, keepaspectratio, trim={27cm 29cm 42cm 3cm}, clip]{pic_ood.pdf}
    
    \vspace{0.5mm}
    \small (b)
\end{minipage}

\vspace{1mm}

\begin{minipage}[t]{0.48\columnwidth}
    \centering
    \includegraphics[height=3.2cm, keepaspectratio, trim={7cm 18cm 65cm 17cm}, clip]{pic_ood.pdf}
    
    \vspace{0.5mm}
    \small (c)
\end{minipage}
\hfill
\begin{minipage}[t]{0.48\columnwidth}
    \centering
    \includegraphics[height=3.2cm, keepaspectratio, trim={31cm 15cm 37cm 16cm}, clip]{pic_ood.pdf}
    
    \vspace{0.5mm}
    \small (d)
\end{minipage}

\vspace{0mm}
\captionsetup{font=normalsize}
\caption{Trajectory visualization of our model under Cross-Object and Cross-Scene generalization settings. Figures (a) and (b) illustrate the model's performance in cluttered scenes, while figures (c) and (d) show its behavior on unseen objects.}
\label{fig:ood}
\vspace{0mm}
\end{figure}

We further evaluate whether the proposed affordance representations can achieve strong zero-shot generalization in real-world manipulation. Specifically, we consider three representative tasks, namely pickup, open, and cut, and assess our model under two settings: cross-object and cross-scene. In each setting, performance is evaluated across all three tasks.

We compare our method with the baselines across both settings and find that it generalizes consistently better in each case. As shown in Table~\ref{tab:generalization_axes}, our method not only bridges the gap between human demonstrations and robot execution, but also remains robust under shifts in object identity and scene background. These results further indicate that affordance-based representations capture transferable interaction structure that supports zero-shot generalization in real-world manipulation.

\subsection{Ablation and Analysis}

In this section, we conduct systematic ablations on the key components and design choices of our method, and quantitatively analyze their contributions. We use 3D Average Displacement Error (ADE) and Final Displacement Error (FDE) in centimeters\cite{bao2023uncertainty, liu2022joint} as evaluation metrics. 

\textbf{Ablation on loss design.} We employ a weighted loss that adaptively reweights supervision according to motion magnitude, such that points with larger displacements receive higher weights while near-static points are down-weighted. To evaluate its effect, we remove this adaptive reweighting while keeping all other settings unchanged. As reported in Table~\ref{tab:ablation}, removing this design leads to degraded performance, confirming the effectiveness of motion-aware reweighting.

\textbf{Effect of the backbone.} For the point-cloud encoder, we compare two widely used backbones, PointNeXt~\cite{qian2022pointnext} and Point Transformer v3~\cite{wu2024point}. Under our training setup, Point Transformer v3 incurs a higher computational cost than PointNeXt, resulting in lower training throughput. Interestingly, PointNeXt achieves slightly better final performance on our benchmarks, which does not fully align with common observations in point-cloud representation learning. Meanwhile, Point Transformer v3 exhibits more stable optimization and faster convergence, with lower variance across random seeds. Both models are trained on the same training set for 2000 epochs.


\begin{table}[t]
\centering
\caption{Ablation study on backbone choice, text fusion strategy, and weighted loss design. Best results in \textbf{bold}. ($\downarrow$ denotes lower is better.)}
\label{tab:ablation}
\renewcommand{\arraystretch}{1.25}
\setlength{\tabcolsep}{6pt}
\small
\begin{tabular}{lccc}
\toprule
Method & ADE($\downarrow$) & FDE($\downarrow$) & Params($\downarrow$) \\
\midrule
PTv3 + Early Fusion        & 0.0596 & 0.0829 & 51.25M \\
PTv3 + Late Fusion         & 0.0445 & 0.0614 & 51.25M \\
PointNeXt + Early Fusion   & 0.0380 & 0.0504 & 8.89M \\
PointNeXt + Late Fusion    & 0.0380 & 0.0504 & 8.89M \\
PointNeXt + Early + WLoss  & 0.0354 & 0.0472 & 8.89M \\
\textbf{PointNeXt + Late + WLoss} & \textbf{0.0350} & \textbf{0.0468} & \textbf{8.89M} \\
\bottomrule
\end{tabular}
\vspace{-10pt}
\end{table}

\textbf{Ablation on text fusion.} We compare two strategies for incorporating language conditioning: early fusion and late fusion. Early fusion injects language features before the point-cloud backbone, whereas late fusion performs the fusion after backbone encoding on the resulting point-cloud features. Empirically, late fusion achieves slightly better performance than early fusion. We attribute this to the fact that point-cloud backbones primarily specialize in geometric modeling and are less suited for direct text-semantic alignment, making high-level feature fusion a more effective way to incorporate language conditions.

\section{Conclusion}
In this paper, we presented BridgeACT, an affordance-driven framework that bridges human demonstrations and robot actions without requiring any robot demonstration data. By modeling affordance as an embodiment-agnostic intermediate representation, BridgeACT unifies manipulation as two complementary problems: where to grasp and how to move. With explicit tool-target interaction modeling, our method enables direct transfer from human videos to real-world robot execution. Experimental results show that BridgeACT outperforms prior baselines on real-world manipulation tasks and generalizes well to unseen objects and scenes. These findings suggest that executable affordance representations provide an effective and scalable interface for learning robot manipulation from large-scale human demonstrations.

\section{Limitation}
A key limitation of BridgeACT is that both rotational motion and fine-grained manipulation remain challenging in human-to-robot transfer. Rotational motion is inherently harder to infer robustly from human demonstrations, while fine-grained manipulation imposes stricter requirements on contact modeling and control precision. As a result, the current method still has limitations in these two scenarios.

\bibliographystyle{IEEEtran}
\bibliography{ref}

@inproceedings{liu2022hoi4d,
  title={Hoi4d: A 4d egocentric dataset for category-level human-object interaction},
  author={Liu, Yunze and Liu, Yun and Jiang, Che and Lyu, Kangbo and Wan, Weikang and Shen, Hao and Liang, Boqiang and Fu, Zhoujie and Wang, He and Yi, Li},
  booktitle={Proceedings of the IEEE/CVF Conference on Computer Vision and Pattern Recognition},
  pages={21013--21022},
  year={2022}
}

@article{damen2020epic,
  title={The epic-kitchens dataset: Collection, challenges and baselines},
  author={Damen, Dima and Doughty, Hazel and Farinella, Giovanni Maria and Fidler, Sanja and Furnari, Antonino and Kazakos, Evangelos and Moltisanti, Davide and Munro, Jonathan and Perrett, Toby and Price, Will and others},
  journal={IEEE Transactions on Pattern Analysis and Machine Intelligence},
  volume={43},
  number={11},
  pages={4125--4141},
  year={2020},
  publisher={IEEE}
}

@inproceedings{damen2018scaling,
  title={Scaling egocentric vision: The epic-kitchens dataset},
  author={Damen, Dima and Doughty, Hazel and Farinella, Giovanni Maria and Fidler, Sanja and Furnari, Antonino and Kazakos, Evangelos and Moltisanti, Davide and Munro, Jonathan and Perrett, Toby and Price, Will and others},
  booktitle={Proceedings of the European conference on computer vision (ECCV)},
  pages={720--736},
  year={2018}
}

@inproceedings{yuan2025roboengine,
  title={Roboengine: Plug-and-play robot data augmentation with semantic robot segmentation and background generation},
  author={Yuan, Chengbo and Joshi, Suraj and Zhu, Shaoting and Su, Hang and Zhao, Hang and Gao, Yang},
  booktitle={2025 IEEE/RSJ International Conference on Intelligent Robots and Systems (IROS)},
  pages={7622--7629},
  year={2025},
  organization={IEEE}
}

@misc{coil,
  title={Correspondence-Oriented Imitation Learning: Flexible Visuomotor Control with 3D Conditioning},
  author={Cao, Yunhao and Bhaumik, Zubin and Jia, Jessie and He, Xingyi and Fang, Kuan},
  year={2025},
  eprint={2512.05953},
  archivePrefix={arXiv},
  primaryClass={cs.RO}
}

@misc{pointworld,
  title={PointWorld: Scaling 3D World Models for In-The-Wild Robotic Manipulation},
  author={Huang, Wenlong and Chao, Yu-Wei and Mousavian, Arsalan and Liu, Ming-Yu and Fox, Dieter and Mo, Kaichun and Fei-Fei, Li},
  year={2026},
  eprint={2601.03782},
  archivePrefix={arXiv},
  primaryClass={cs.RO}
}

@misc{tracegen,
  title={TraceGen: World Modeling in 3D Trace Space Enables Learning from Cross-Embodiment Videos},
  author={Lee, Seungjae and Jung, Yoonkyo and Chun, Inkook and Lee, Yao-Chih and Cai, Zikui and Huang, Hongjia and Talreja, Aayush and Dao, Tan Dat and Liang, Yongyuan and Huang, Jia-Bin and Huang, Furong},
  year={2025},
  eprint={2511.21690},
  archivePrefix={arXiv},
  primaryClass={cs.RO}
}

@article{myers2024policy,
  title={Policy adaptation via language optimization: Decomposing tasks for few-shot imitation},
  author={Myers, Vivek and Zheng, Bill Chunyuan and Mees, Oier and Levine, Sergey and Fang, Kuan},
  journal={arXiv preprint arXiv:2408.16228},
  year={2024}
}

@inproceedings{bharadhwaj2024track2act,
  title={Track2act: Predicting point tracks from internet videos enables generalizable robot manipulation},
  author={Bharadhwaj, Homanga and Mottaghi, Roozbeh and Gupta, Abhinav and Tulsiani, Shubham},
  booktitle={European Conference on Computer Vision},
  pages={306--324},
  year={2024},
  organization={Springer}
}

@article{brohan2022rt,
  title={Rt-1: Robotics transformer for real-world control at scale},
  author={Brohan, Anthony and Brown, Noah and Carbajal, Justice and Chebotar, Yevgen and Dabis, Joseph and Finn, Chelsea and Gopalakrishnan, Keerthana and Hausman, Karol and Herzog, Alex and Hsu, Jasmine and others},
  journal={arXiv preprint arXiv:2212.06817},
  year={2022}
}

@article{wen2023any,
  title={Any-point trajectory modeling for policy learning},
  author={Wen, Chuan and Lin, Xingyu and So, John and Chen, Kai and Dou, Qi and Gao, Yang and Abbeel, Pieter},
  journal={arXiv preprint arXiv:2401.00025},
  year={2023}
}

@article{niu2025pre,
  title={Pre-training auto-regressive robotic models with 4d representations},
  author={Niu, Dantong and Sharma, Yuvan and Xue, Haoru and Biamby, Giscard and Zhang, Junyi and Ji, Ziteng and Darrell, Trevor and Herzig, Roei},
  journal={arXiv preprint arXiv:2502.13142},
  year={2025}
}

@inproceedings{karaev2025cotracker3,
  title={Cotracker3: Simpler and better point tracking by pseudo-labelling real videos},
  author={Karaev, Nikita and Makarov, Yuri and Wang, Jianyuan and Neverova, Natalia and Vedaldi, Andrea and Rupprecht, Christian},
  booktitle={Proceedings of the IEEE/CVF International Conference on Computer Vision},
  pages={6013--6022},
  year={2025}
}

@article{carion2025sam,
  title={Sam 3: Segment anything with concepts},
  author={Carion, Nicolas and Gustafson, Laura and Hu, Yuan-Ting and Debnath, Shoubhik and Hu, Ronghang and Suris, Didac and Ryali, Chaitanya and Alwala, Kalyan Vasudev and Khedr, Haitham and Huang, Andrew and others},
  journal={arXiv preprint arXiv:2511.16719},
  year={2025}
}

@inproceedings{wang2025vlm,
  title={Vlm see, robot do: Human demo video to robot action plan via vision language model},
  author={Wang, Beichen and Zhang, Juexiao and Dong, Shuwen and Fang, Irving and Feng, Chen},
  booktitle={2025 IEEE/RSJ International Conference on Intelligent Robots and Systems (IROS)},
  pages={17215--17222},
  year={2025},
  organization={IEEE}
}

@article{huang2024rekep,
  title={Rekep: Spatio-temporal reasoning of relational keypoint constraints for robotic manipulation},
  author={Huang, Wenlong and Wang, Chen and Li, Yunzhu and Zhang, Ruohan and Fei-Fei, Li},
  journal={arXiv preprint arXiv:2409.01652},
  year={2024}
}

@article{intelligence2025pi_,
  title= {$\pi_{0.5}$: a Vision-Language-Action Model with Open-World Generalization},
  author={Intelligence, Physical and Black, Kevin and Brown, Noah and Darpinian, James and Dhabalia, Karan and Driess, Danny and Esmail, Adnan and Equi, Michael and Finn, Chelsea and Fusai, Niccolo and others},
  journal={arXiv preprint arXiv:2504.16054},
  year={2025}
}

@inproceedings{tang2025uad,
  title={Uad: Unsupervised affordance distillation for generalization in robotic manipulation},
  author={Tang, Yihe and Huang, Wenlong and Wang, Yingke and Li, Chengshu and Yuan, Roy and Zhang, Ruohan and Wu, Jiajun and Fei-Fei, Li},
  booktitle={2025 IEEE International Conference on Robotics and Automation (ICRA)},
  pages={3822--3831},
  year={2025},
  organization={IEEE}
}

@article{kim2024openvla,
  title={Openvla: An open-source vision-language-action model},
  author={Kim, Moo Jin and Pertsch, Karl and Karamcheti, Siddharth and Xiao, Ted and Balakrishna, Ashwin and Nair, Suraj and Rafailov, Rafael and Foster, Ethan and Lam, Grace and Sanketi, Pannag and others},
  journal={arXiv preprint arXiv:2406.09246},
  year={2024}
}

@article{yuan2024general,
  title={General flow as foundation affordance for scalable robot learning},
  author={Yuan, Chengbo and Wen, Chuan and Zhang, Tong and Gao, Yang},
  journal={arXiv preprint arXiv:2401.11439},
  year={2024}
}

@article{xu2024flow,
  title={Flow as the cross-domain manipulation interface},
  author={Xu, Mengda and Xu, Zhenjia and Xu, Yinghao and Chi, Cheng and Wetzstein, Gordon and Veloso, Manuela and Song, Shuran},
  journal={arXiv preprint arXiv:2407.15208},
  year={2024}
}

@inproceedings{wu2024point,
  title={Point transformer v3: Simpler faster stronger},
  author={Wu, Xiaoyang and Jiang, Li and Wang, Peng-Shuai and Liu, Zhijian and Liu, Xihui and Qiao, Yu and Ouyang, Wanli and He, Tong and Zhao, Hengshuang},
  booktitle={Proceedings of the IEEE/CVF conference on computer vision and pattern recognition},
  pages={4840--4851},
  year={2024}
}

@article{qian2022pointnext,
  title={Pointnext: Revisiting pointnet++ with improved training and scaling strategies},
  author={Qian, Guocheng and Li, Yuchen and Peng, Houwen and Mai, Jinjie and Hammoud, Hasan and Elhoseiny, Mohamed and Ghanem, Bernard},
  journal={Advances in neural information processing systems},
  volume={35},
  pages={23192--23204},
  year={2022}
}

@article{bai2025qwen3,
  title={Qwen3-vl technical report},
  author={Bai, Shuai and Cai, Yuxuan and Chen, Ruizhe and Chen, Keqin and Chen, Xionghui and Cheng, Zesen and Deng, Lianghao and Ding, Wei and Gao, Chang and Ge, Chunjiang and others},
  journal={arXiv preprint arXiv:2511.21631},
  year={2025}
}

@inproceedings{radford2021learning,
  title={Learning transferable visual models from natural language supervision},
  author={Radford, Alec and Kim, Jong Wook and Hallacy, Chris and Ramesh, Aditya and Goh, Gabriel and Agarwal, Sandhini and Sastry, Girish and Askell, Amanda and Mishkin, Pamela and Clark, Jack and others},
  booktitle={International conference on machine learning},
  pages={8748--8763},
  year={2021},
  organization={PmLR}
}

@article{vaswani2017attention,
  title={Attention is all you need},
  author={Vaswani, Ashish and Shazeer, Noam and Parmar, Niki and Uszkoreit, Jakob and Jones, Llion and Gomez, Aidan N and Kaiser, {\L}ukasz and Polosukhin, Illia},
  journal={Advances in neural information processing systems},
  volume={30},
  year={2017}
}

@inproceedings{bao2023uncertainty,
  title={Uncertainty-aware state space transformer for egocentric 3d hand trajectory forecasting},
  author={Bao, Wentao and Chen, Lele and Zeng, Libing and Li, Zhong and Xu, Yi and Yuan, Junsong and Kong, Yu},
  booktitle={Proceedings of the IEEE/CVF international conference on computer vision},
  pages={13702--13711},
  year={2023}
}

@inproceedings{liu2022joint,
  title={Joint hand motion and interaction hotspots prediction from egocentric videos},
  author={Liu, Shaowei and Tripathi, Subarna and Majumdar, Somdeb and Wang, Xiaolong},
  booktitle={Proceedings of the IEEE/CVF conference on computer vision and pattern recognition},
  pages={3282--3292},
  year={2022}
}

@inproceedings{hang2023efficient,
  title={Efficient diffusion training via min-snr weighting strategy},
  author={Hang, Tiankai and Gu, Shuyang and Li, Chen and Bao, Jianmin and Chen, Dong and Hu, Han and Geng, Xin and Guo, Baining},
  booktitle={Proceedings of the IEEE/CVF international conference on computer vision},
  pages={7441--7451},
  year={2023}
}

\end{document}